%% file: main.tex
\documentclass{article}

\usepackage{microtype}
\usepackage{graphicx}
\usepackage{subfigure}
\usepackage{amsmath}
\usepackage{amsfonts}
\usepackage{marginnote}
\usepackage{booktabs}
\usepackage{makecell}
\usepackage{multirow}
\usepackage{bm}
\usepackage{float}

\usepackage[dvipsnames]{xcolor}

\usepackage{qtree}
\usepackage{tikz}
\usepackage{tikz-qtree}
\usepackage{lscape}

\usepackage{hyperref}

\usepackage[accepted]{hill2019}

\icmltitlerunning{Concept Tree: High-Level Representation of Variables for More Interpretable Surrogate Decision Trees}

\begin{document}

\twocolumn[
\icmltitle{Concept Tree: High-Level Representation of Variables for More Interpretable Surrogate Decision Trees}



\icmlsetsymbol{equal}{*}

\begin{icmlauthorlist}
\icmlauthor{Xavier Renard}{equal,axa}
\icmlauthor{Nicolas Woloszko}{equal,oecd}
\icmlauthor{Jonathan Aigrain}{equal,axa}
\icmlauthor{Marcin Detyniecki}{axa,lip6,polish}
\end{icmlauthorlist}

\icmlaffiliation{axa}{AXA, Paris, France}
\icmlaffiliation{oecd}{OECD, Paris, France}
\icmlaffiliation{lip6}{Sorbonne Universite, CNRS, LIP6, ´
Paris, France}
\icmlaffiliation{polish}{Polish Academy of
Science, IBS PAN, Warsaw, Poland}

\icmlcorrespondingauthor{Xavier Renard}{xavier.renard@axa.com}
\icmlcorrespondingauthor{Nicolas Woloszko}{nicolas.woloszko@oecd.org}

\icmlkeywords{interpretability, transparency}

\vskip 0.3in
]



\printAffiliationsAndNotice{\icmlEqualContribution} 

\begin{abstract}
Interpretable surrogates of black-box predictors trained on high-dimensional tabular datasets can struggle to generate comprehensible explanations in the presence of correlated variables.
We propose a model-agnostic interpretable surrogate that provides global and local explanations of black-box classifiers to address this issue.
We introduce the idea of \emph{concepts} as intuitive groupings of variables that are either defined by a domain expert or automatically discovered using correlation coefficients. Concepts are embedded in a surrogate decision tree to enhance its comprehensibility.
First experiments on FRED-MD, a macroeconomic database with 134 variables, show improvement in human-interpretability while accuracy and fidelity of the surrogate model are preserved.
\end{abstract}

\section{Introduction}

The field of interpretability aims at providing users and practitioners with techniques meant to explain either globally a trained machine learning model or locally a particular prediction made by a model. This can be achieved either by training directly an interpretable model, or in a post hoc approach, using model-agnostic or model-specific interpretability techniques.

This paper focuses on post hoc surrogate models that globally approximate a machine learning classifier while providing explanations at the local level of each prediction.
We are interested in model-agnostic interpretability approaches meant to be applied on standard feature spaces composed of tabular data. Our goal is to explain any type of trained model: the classifier is a black-box left to the discretion of the practitioners.
We refer the reader to recent published surveys for a global picture of the interpretability field as for instance~\citep{Guidotti2018a}.

\begin{figure}
    \centering
    \resizebox*{\columnwidth}{3.5cm}{
    \input{illustration_tree_concept.tex}
    }
    \caption{Concept Tree trained on FRED-MD macroeconomic dataset. Variables are grouped by Concepts to constraint the training of an interpretable surrogate decision tree}
    \label{fig:my_label}
\end{figure}
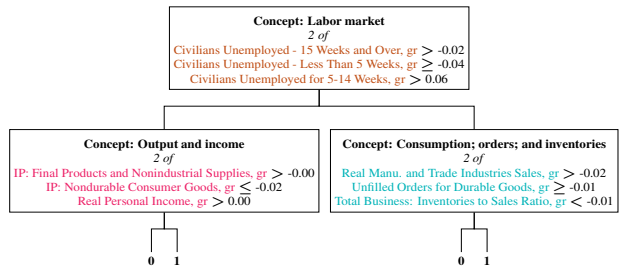

Surrogate models aiming at providing post hoc interpretability may induce confusion by conveying a false sense of simplicity, especially when subgroups of dependent variables are involved. We refer to dependent variables as variables sharing similar information and possibly generated by a common phenomenon. It may include the various lags of a given time series, various features of a variables, or various measures of a given fact. Surrogate models may arbitrarily select one given variable among a group of dependent variables, thus obscuring the global picture.
Subsequently, practitioners may better understand a surrogate model that retains the whole set of dependent variables and depicts a bigger picture than a simpler model. 

This paper introduces the idea of \textit{concept}. A \textit{concept} is a representation gathering a group of dependent variables. It can be defined using either domain knowledge or statistical properties of dependent variables (such as the Pearson correlation). The use of \textit{concepts} allows to provide high-level representations that practitioners may find easier to interpret. We contend that \textit{concept}-based methods may be better suited to human understanding and provide more practitioner-friendly representations of a black-box classifier. 

We substantiate that claim with an application to decision tree surrogates. Decision trees are universally considered interpretable by domain experts~\cite{freitas2014comprehensible}. We compare standard surrogate tree models to trees whose training is constrained by the grouping of subgroups of variables in \textit{concepts}. More specifically, we embed the idea of \textit{concept} in the TREPAN algorithm~\citep{Craven1996b}, an interpretable decision tree originally instantiating a variant of \emph{id2-of-3}~\citep{murphy1991id2} with a mechanism of oracle querying aiming at populating areas of the training set where the fidelity of the surrogate can be improved. In our approach, the \textit{concepts} are used at each node of the decision tree to constrain the training of the split rule based on \emph{id2-of-3}. We compare the resulting \textit{Concept Trees} to the surrogates provided by the original TREPAN algorithm. 

The next section expands on the motivation and formally introduces the idea of \textit{high-level concepts}. Section 3 introduces \textit{Concept Trees}, a version of the TREPAN algorithm that builds on \textit{concepts}, and shows that \textit{Concept Trees} meet the prerequisites of a global-to-local, post-hoc and model-agnostic surrogate. Section 4 assesses both the qualitative and quantitative relevance of our proposition through experiments led on FRED-MD, a monthly macroeconomic database designed for empirical analysis of the US economy \citep{mccracken2016fred}.

\section{Concept: Grouping Dependent Variables into High-Level Representation of Variables}

It is often the case that groupings of variables in a given dataset may naturally appear. Such grouping can derive from similar meaning or a similar origin (\emph{e.g.} unemployment among men, unemployment among women, unemployment among young people...). A grouping can also be the result of multiple transformations applied to a given source of data (such as multiple lags of a time series, or features engineered from the same variable).

In this work, we consider two types of \textit{concepts}: expert-defined grouping of features and automatically-defined grouping based on a statistical criterion such as feature correlation. 
Expert-based \textit{concepts} may be used when domain knowledge is available. Automatically-defined concepts do not require prior domain knowledge.

Exploiting the group structure of variables has already been used in the literature to train more accurate sparse models, for instance with \emph{group-lasso}~\citep{yuan2006model} or \emph{sparse-group-lasso}~\citep{simon2013sparse}. In the latter, improved accuracy is observed with variable groupings such as gene pathways or factor level indicators in categorical data.
Other machine learning fields also cover the idea of grouping dimensions, such as subspace clustering \citep{vidal2011subspace}.

In the field of interpretability, the idea of exploiting a meaningful grouping of features to generate better explanations has emerged, for instance with topic-modeling-based feature compression~\citep{kim2015scalable} or on image classification with deep learning models~\citep{Kim2017,Ghorbani2019}.

Correlated features is a known challenge when building machine learning models and interpreting feature importances~\cite{Buhlmann2013,Gregorutti2017,Strobl2008,Tolosi2011}. For instance, \emph{lasso}-based methods for feature selection tend to select only one representative from a group of correlated features and to discard the others~\cite{Buhlmann2013}. It has been pointed out that correlated features severely impact variable importance measures of random forests~\cite{Strobl2008, Gregorutti2017}. Also, many feature selection methods suffer from a \emph{correlation bias}: features belonging to a group of correlated features receive weights inversely proportional to the size of the group~\cite{Tolosi2011}. This issue creates instability in the feature selection process. Small changes in the training data can result in significant changes in the selected set of features. This instability prevents a robust interpretation of variable importance.

We propose to use the idea of \textit{concept} to address both expert-defined grouping of features and automatically (correlated)-defined grouping. \textit{Concepts} are embedded into surrogate models in order to constrain their training, which provides two levels of granularity for the explanations: at high-level (concept) and at finer level (raw variables).
The next paragraph offers a formal presentation of the idea of \textit{concepts}. 

We consider a set of training examples $\mathbb{X}$ where each example is denoted $x^{(i)}$ with $i \in [1...|\mathbb{X}|]$ and associated with a label $y^{(i)}$. The set of training examples $\mathbb{X}$ is composed of a set of features $j \in \mathbb{J}$ and each feature vector is noted  $x_j$ with $j \in \mathbb{J}=[1...N]$.

A \emph{concept} is a subset of features denoted $c_k \subset \mathbb{J}$. $K$ concepts $c_k$ co-exist to form the set of concepts $c_k \in \mathbb{C}, k \in [1...K]$. The instantiation of a concept $c_k$ is the process of populating it with dependent features. Every feature $j \in \mathbb{J}$ belongs to a single concept $c_k$ and one concept only:
$$c_k \cap c_l = \emptyset \mid \forall l \in \mathbb{J} \text{ and } l \neq k$$

\subsection{Expert knowledge concepts}
The instantiation of a concept $c_k$ can be either driven by domain knowledge or performed automatically. The former requires that all variables belong to user-defined groups that be meaningful to domain experts. The variable classifications are sometimes to be found in the documentation of a dataset. That is the case of the FRED-MD data, which is used in the experimentation section of this work. The paper accompanying the dataset \citep{mccracken2016fred} includes in appendix a table that classifies the 134 monthly macroeconomic indicators into 8 categories: output and income, labour market, housing, consumption orders and inventories, money and credit, interest and exchange rates, prices, and stock markets. Table~\ref{tab:fred_succint_desc} provides a sample of these categories. 

\begin{table*}[]
\centering
\caption{Overview of the grouping of variables by concept in FRED-MD database~\citep{mccracken2016fred}}
\label{tab:fred_succint_desc}
\resizebox{\textwidth}{!}{%
\begin{tabular}{@{}ccc@{}}
\toprule
\thead{Concept 1: Output and Income}     & \thead{Concept 2: Labor Market} & \thead{Concept 5: Money and Credit}        \\ \midrule
Real Personal Income                      & Civilian Labor Force             & Total Reserves of Depository Institutions   \\
Real personal income ex transfer receipts & Civilian Employment              & Commercial and Industrial Loans             \\
IP: Consumer Goods                        & Civilian Unemployment Rate       & Total Consumer Loans and Leases Outstanding \\
...                                       & ...                              & ...                                         \\ \bottomrule
\end{tabular}
}
\end{table*}

\subsection{Automatic concepts: a simple approach}
Failing that user may rely on domain knowledge, the set of \textit{concepts} $\mathbb{C}$ can be built automatically using a clustering algorithm based on feature correlations. Features (indexed by $j$) can be grouped in a concept $c_k$ using any dependence measure $\rho$. The most straightforward is the Pearson correlation, that measures linear correlation between variables. Assuming the measure has values between $[-1;1]$ (an absolute value of $1$ meaning two variables perfectly dependent), a user-defined threshold $\epsilon$ is set on the absolute value of the measure of dependence between two features $x_j$ and $x_{j'}$ in order to decide whether these features belong to the same concept $c_k$:

$$\left|\rho(x_j,x_{j'})\right| \geq \epsilon \mid \forall (j, j') \in c_k$$

The clustering algorithm is greedy: for each iteration a feature is tested against all features and existing groups. A feature $j'$ is affected to a \textit{concept} $c_k$ if its dependence to each feature in $c_k$ is higher than $\epsilon$:
$$\left|\rho(x_j,x_{j'})\right| \geq \epsilon \mid \forall j \in c_k \rightarrow c_k = c_k \cup j'$$

If a given feature is independent from all the others, it belongs to a singleton.
This formalization is also adequate for the expert's knowledge grouping. In that case, $\rho$ and $\epsilon$ would be the criteria of group assignment by the expert.

The next section explains how the notion of  \textit{concepts} may be used to constrain the training of a decision tree in order to produce more interpretable surrogates.

\section{Concept Tree: Embedding Concepts For More Interpretable Surrogate Decision Tree}

Decision trees are a well-known interpretable machine learning model. A decision tree has a graphical structure, its decisions rely on a sparse subset of features, and features are used in a hierarchical way, thus conveying an intuitive sense of feature importance and providing several levels of explanation granularity~\citep{freitas2014comprehensible}. Training a decision tree on the training set $\mathbb{X}$ yields an interpretable classification algorithm, provided that the number of nodes is kept under a certain threshold. The limit on the tree complexity may come at the expense of predictive performance. Decision trees appear as good candidate surrogates to black-box classifiers. 

A decision tree surrogate is produced as follows. A black-box $b$ is trained on $\mathbb{X}$ with the true class labels $y^{(i)}$ ; the surrogate $f$ is then trained on the black-box predictions $\hat{y}^{(i)} = b(x^{(i)})$. In production, the classification is performed by the black-box while the explanations are provided by the surrogate decision tree. The fidelity of the surrogate is assessed as the proportion of instances where the surrogate makes the same prediction than the black-box classifier. 


The TREPAN algorithm is an instance of interpretable surrogate tree model~\citep{Craven1996}. It is model-agnostic and aims at mimicking the classification behaviour of a black-box $b$. It queries the black-box with instances to get predictions $\hat{y}^{(i)} = b(x^{(i)})$ and then fits an interpretable decision tree. The outline of TREPAN is shown in Algorithm~\ref{alg:trepan}. The querying of extra instances allows to populate the critical areas of the feature space and thus significantly curb the tendency of decision trees to overfit. 


TREPAN uses $m-of-n$ decision rules, that are inspired from $id2-of3$ decision trees~\citep{murphy1991id2}. To fit an $m-of-n$ decision rule, the set of the $n$ most discriminative tests on the features for the node is discovered using the information gain. Then, in order to validate a node, an instance must validate at least $m$ tests among the $n$. For instance, given a decision rule with 3 tests $x_1$, $x_2$ and $x_3$, such as $2$-of-$\{x_1, \neg x_2, x_3\}$ is equivalent to the logical expression $(x_1 \vee \neg x_2) \wedge (x_1 \vee x_3) \wedge (\neg x_2 \vee x_3)$.
The parameters $m$ and $n$ are user-defined upper-bounds: their final values are learnt by the node. The $m-of-n$ decision rules are learnt in a greedy way for computational efficiency. For the outline of the fitting algorithm of an $m-of-n$ decision rule, we refer the reader to the original paper~\citep{Craven1996b} for the sake of conciseness and precision.

While the original TREPAN paper is two decades old already, researchers have kept reassessing its relevance up until recently~\citep{Sarkar2016}.
Experimentations show that TREPAN has a good fidelity to the black-box and a better accuracy on the test set than a decision tree directly trained on the training set $\mathbb{X}$~\citep{Craven1996}. This good performance is attributed to the additional-instance-drawing mechanism, which yields a denser support to the fit of a decision rule and thus a better prediction accuracy. 

The $m-of-n$ decision rule structure improves the accuracy and the fidelity of the decision tree as it allows to learn more complex decision boundaries. However, it comes at the price of interpretability of both the node's decision rule and the decision tree overall. A practitioner may find it hard to understand all the possible ${n \choose m}$ combinations of variables at the same time. Moreover, the contrary of a $m-of-n$ literal may be challenging to conceive as soon as $m>1$ and $1<n<m$. Alternatively, simpler nodes such as the ones in $C4.5$~\citep{quinlan1993c4} would be easier to interpret, but would yield significantly larger trees for a given fidelity level, thus reducing overall interpretability ~\citep{Craven1996}.

Similar to TREPAN, the Concept Tree aims at mimicking the classification behaviour of $b$, using the additional-instance-drawing procedure at the node level to improve fidelity. The chief difference between Concept Tree and TREPAN lies in the learning of the decision rule at each node. The $m-of-n$ decision rule is no longer fitted on the whole feature space. It is constrained to use subsets of features defined by the \emph{concepts}. At a given node, the Concept Tree fits a $m-of-n$ decision rule using only the variables related to the \textit{concept} $c_k$ for each \textit{concept} $c_k$, and selects the one that yields the best information gain. Each node thus splits the sample on a $m-of-n$ decision rule based on a \emph{concept}, using related variables only. A part from that restriction, the Concept Tree uses the same expanding procedure as TREPAN, described in Algorithm~\ref{alg:trepan}.

This paper aims at improving the interpretability of surrogate trees built with $m-of-n$ decision rules by introducing the Concept Tree, a tree-based surrogate methods based on TREPAN and the use of \textit{concepts}. 
The use of \textit{concepts} is expected to help practitioners better understand the surrogate. Each node relies on variables belonging to one \textit{concept}-grouping only. Nodes thus use complex $m-of-n$ literals but ensure better human-understandability by organizing information at a \textit{concept} level.

We argue that concept-based decision rules have a better interpretability than standard $m-of-n$ decision rules while having the exact same informational complexity (the number of bits needed to write down the decision rule). Defying the conventional notion of complexity-interpretablity trade-off, Concept Trees achieve a higher interpretability at the same level of complexity, thus preserving predictive accuracy.

\begin{algorithm}[tb]
   \caption{Simplified overview of Trepan}
   \label{alg:trepan}
\begin{algorithmic}
    \STATE {\bfseries Trepan}($b$, $X$,  $max\_nodes$, $min\_sample$, $m$, $n$)
\STATE Initialize the tree with root $R$
\STATE $S \gets X \cup DrawSample(min\_sample-|X|)$
\STATE Get labels from black-box $b$ for train set $S$
\STATE Initialize $Queue$ with $<R, S>$
\STATE $n\_nodes = 1$
\WHILE{$Queue \neq \emptyset$ and $n\_nodes < max\_nodes$}
\STATE Remove $<node N, S_N>$ from head of $Queue$ 
\STATE Fit decision rule of node $N$
\FOR{each outcome $t$ of the test}
    \STATE Initialize a child node $C$
    \STATE $S_c \gets$ instances of $S_N$ with outcome $t$ for the test
    \STATE $S_C \gets S_c \cup DrawSample(min\_sample-|S_c|)$
    \STATE Get labels from black-box $b$ for $S_C$
    \IF{$C$ is not pure enough}
        \STATE Add $<node C, S_C>$ to $Queue$
    \ENDIF
    \STATE $n\_nodes = n\_nodes + 1$
\ENDFOR
\ENDWHILE

\STATE {\bfseries Return} $R$

\end{algorithmic}
\end{algorithm}

\begin{algorithm}[tb]
   \caption{Construction of a Concept Tree decision rule}
   \label{alg:concepttreenode}
\begin{algorithmic}
   \STATE {\bfseries ConstructConceptDecisionRule}($X$, $y$, $concepts$)
   
   \STATE $best\_candidate \leftarrow \emptyset$
   \STATE $best\_ig \leftarrow = 0$
   \FOR{$c \in concepts$}
   \STATE $X_c \leftarrow$ Select features from $X$ belonging to $c$
   \STATE $candidate \leftarrow MofNDecisionRule(X_c, y, m, n)$
   \STATE $ig \leftarrow$ Compute information gain for $candidate$
   \IF{$ig > best\_ig$}
    \STATE $best\_ig \leftarrow ig$
    \STATE $best\_candidate \leftarrow candidate$
   \ENDIF
   \ENDFOR
   
   \STATE {\bfseries Return} $best\_candidate$
 
\end{algorithmic}
\end{algorithm}

\section{Experimentation: FRED-MD Macroeconomic Database}

This paper has introduced the ideas of Concept and Concept Tree, whose main objectives are to provide an accurate surrogate $f$ mimicking a black-box classifier $b$ while being as interpretable as possible. The next paragraphs describe experimentations made with the FRED-MD dataset ~\citep{mccracken2016fred}, a publicly released macroeconomic database of 134 monthly U.S. indicators with more than 700 instances. Interpretability is critical in economics and our experimentations show how Concept Trees may match the requirements of the field. 

The experimentations are conduced as follows. The classification target is computed from the \emph{civilian unemployment rate}: if the value for an instance is lower than in the previous period, the target value is set to label 0 and the label is set to 1 otherwise. Domain-knowledge-based experts are extracted from the FRED-MD official documentation, which classifies variables into 8 subgroups (see Table~\ref{tab:fred_succint_desc}). 

The competitors are both flavors of Concept Tree (Concept Tree-Expert and Concept Tree-Correlated for automatically-defined concepts) and the original TREPAN.
Since the Concept Tree and TREPAN have a similar structure, they share the same parameters for the experimentation. The maximal number of nodes $max\_nodes$ is set to 10. For the split rules, the values of $m-of-n$ are set to $1-of-1$, $3-of-3$ and $5-of-5$. The minimal value of samples $min\_samples$ to fit a split rule at a node is 100, thus additional samples are drawn from the fitted distribution if the $\mathbb{X}$ is not large enough. For Concept Tree-Correlation, the threshold $\epsilon$ on the correlation $\rho$ is set to $0.9$ such as $\left|\rho(x_j,x_{j'})\right| \geq 0.9$.

The black-box $b$ used is a Random Forest with 200 estimators, with the scikit-learn default values for the other parameters.
Out-sample-fidelity is computed by 5-fold cross-validation. At each split the black-box is fitted on the train set and makes predictions for the train set and the test set. The Concept Tree and TREPAN instances are then fitted on the train set with black-box predictions as targets, and their fidelities are measured against the black-box predictions made on the test set. Out-of-sample accuracy is assessed using the same procedure. Fidelity measures the proportion of predictions made by the surrogate that match the predictions made by the black-box, while accuracy measures the proportion of predictions made by the surrogate that match the actual value of the target. Interpretability is assessed by economic expert judgement.

\begin{table}[!htbp]
\centering
\caption{Experimental results: surrogate accuracy and fidelity as a function of the algorithm, the concept type and the split rule}
\label{tab:results}
\resizebox{\columnwidth}{!}{%
\begin{tabular}{@{}ccc||ccc@{}}
\toprule
\thead{Algorithm}   & \thead{Concept Type} & \thead{Split Rule}  & \thead{Surr. Accuracy} & \thead{Surr. Fidelity} \\ \midrule \midrule
Concept Tree & Expert & \multirow{ 3}{*}{$1-of-1$}  & $63\% \pm 4\%$ & $65\% \pm 9\%$ \\
Concept Tree & Correlation &  & $68\% \pm 6\%$ & $71\% \pm 6\%$ \\
TREPAN & / &  & $\bm{75\% \pm 9\%}$ & $\bm{74\% \pm 7\%}$ \\ \hline
Concept Tree & Expert & \multirow{ 3}{*}{$3-of-3$}  & $69\% \pm 9\%$ & $\bm{76\% \pm 4\%}$ \\
Concept Tree & Correlation &  & $\bm{72\% \pm 11\%}$ & $75\% \pm 5\%$ \\
TREPAN & / &  & $68\% \pm 8\%$ & $72\% \pm 6\%$ \\ \hline
Concept Tree & Expert & \multirow{ 3}{*}{$5-of-5$} & $\bm{71\% \pm 4\%}$ & $\bm{73\% \pm 2\%}$ \\
Concept Tree & Correlation &  & $70\% \pm 8\%$ & $71\% \pm 8\%$ \\
TREPAN & / &  & $67\% \pm 5\%$ & $71\% \pm 4\%$ \\
\bottomrule
\end{tabular}%
}
\end{table}

\subsection{Results}
Table~\ref{tab:results} presents the cross-validated accuracies and fidelities for TREPAN, the Concept Tree with expert-defined concepts and the Concept Tree with automatically defined clusters. The black-box mean accuracy over the folds is $82\% \pm 4\%$.

The experimentations show that Concept Tree provides surrogates whose fidelity and accuracy matches the performance of TREPAN trees and whose interpretability may be significantly enhanced. Although, TREPAN leads in terms of accuracy and fidelity for $1-of-1$ nodes; Concept Tree-Expert for $3-of-3$; and Concept Tree-Correlated for $5-of-5$ nodes, the non-negligible standard-deviations and the setup of this preliminary experiment (number of folds and datasets) don't allow for a final conclusion. However, the experiment highlights the relevance of Concept Tree in terms of accuracy and fidelity and as things stand, Concept Tree is at least as relevant as TREPAN.

Figure \ref{fig:trees} in Appendix displays the trees generated by TREPAN (Figure \ref{fig:tree_trepan}), the Concept Tree with expert-defined concepts (Figure \ref{fig:concept_expert}) and the Concept Tree with correlation-based defined concepts (Figure \ref{fig:concept_correlated}). 

From a macroeconomic point of view, the Concept Tree yields meaningful high level explanations of the workings of the black-box classifiers. The Concept Tree-Expert highlights that Labor Market related variables are the most important in the prediction of the target, followed by Output and Income related variables and Consumption related variables. The Concept Tree-Correlated also sheds light on the importance of nodes referring to Labour market data. Overall, Concept Tree enhances the interpretability of surrogate trees by \textbf{structuring} the explanations.

In Concept Tree-Expert (based on domain-knowledge), explanations are structured by sharing a common "language" with users or experts. It provides the big picture with one general \textit{concept} by node. The detailed analysis of a node is eased because only related, homogeneous, variables are assembled. There is an intuitive relations between high-level explanations (concepts) and low-level explanations (raw variables).

In Concept Tree-Correlation, computed automatically based on variable correlations, part of the domain-knowledge can be recovered. Concept Tree-Correlation presents also the advantage of gathering dependent variables for each node, avoiding arbitrary choices between correlated variables to build a test.

In contrast, TREPAN trees use an idiosyncratic language not shared by the practitioner. Associations of tests in a TREPAN node generate confusion by gathering variables that are hardly related from a domain-knowledge point of view. Such nodes obstruct the understanding by preventing the user from getting the big picture.
    
To illustrate these arguments, we focus on the top three nodes of the trees in Figures \ref{fig:tree_trepan}, \ref{fig:concept_expert} and \ref{fig:concept_correlated}. In the TREPAN tree (Figure \ref{fig:tree_trepan}), the colored variables names highlight that 8 out of the 9 chosen variables are part of the Labor Market concept. This structure is explicitly displayed by the Concept Tree-Expert (Figure \ref{fig:concept_expert}) as the concept chosen in the root node, facilitating the interpretation of the tree by referring to high level concept. We can also notice in the TREPAN tree that the left child of the root node chose the \textit{Civilian Employment, gr} feature for its first rule, whereas the right child node chose the \textit{All Employees: Total nonfarm, gr} feature instead. However, the cluster 3 in the Concept Tree-Correlated (Figure \ref{fig:concept_correlated}) explicitly shows that these features are highly correlated, suggesting that they are interchangeable.



\section{Conclusion}

The present paper introduces \textit{concepts}, a meaningful manner to group dependent variables, and Concept Trees, an alternative tree-based surrogate model that provides both high-level and detailed explanation to black-box classifiers. The grouping of variables in \textit{concepts} allows to overcome the false sense of simplicity conveyed by simpler decision tree surrogate that may give an artificially high importance to a given variable picked among a set of correlated variables, thus obscuring the bigger picture. The use of \textit{concepts} also helps practitioners make sense of otherwise cryptic $m-of-n$ literals, by relying on a higher-level representation of the data. Compared to TREPAN, Concept Trees produce surrogates that have a comparable size and are as accurate, but more easily understandable to a human thanks to a better organization of the information along higher-level representations that significantly enhance the interpretability of the surrogate.
Experiments were conduced using FRED-MD, a macroeconomic database whose documentation includes a grouping of variables. The Concept Tree was applied to this data using both expert-defined concepts derived from the data documentations and concepts built using a simple correlation-based clustering algorithm. First results show a notable improvement in human-readability while accuracy and fidelity of the surrogate are preserved.
Further research could involve a deeper assessment of our propositions, both quantitatively and qualitatively. It could also be relevant to explore alternative clustering algorithms designed to produce more relevant \textit{concepts}.
Further modification to the Concept Tree algorithm may improve performance: currently, following TREPAN's principle, one concept can only be used once in a decision path. Considering a concept encompass several variables, the accuracy and fidelity of the surrogate may suffer from this probably too severe constraint.


\bibliographystyle{icml2019}
\bibliography{TREPAN_intro}

\clearpage

\appendix

\section{Appendix: Decision Trees}

\begin{figure}[H] 
\centering
\resizebox*{16cm}{12cm}{
\subfigure[\textbf{Trepan Decision Tree. Variables from the same expert-defined concepts are displayed with the same color. We can thus easily see that most nodes use variables from heterogeneous groups, making the interpretation difficult}]{ 
\input{tree_trepan.tex}
\label{fig:tree_trepan}
}}

\end{figure}

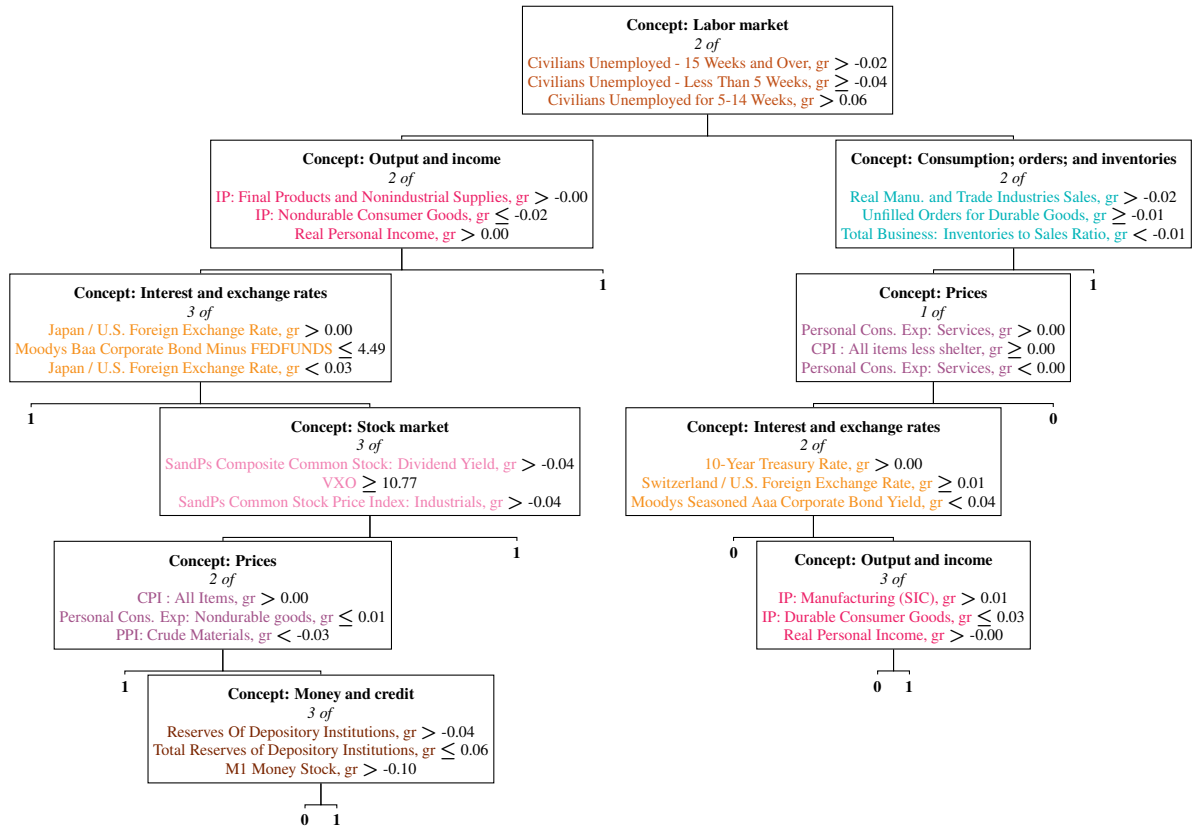
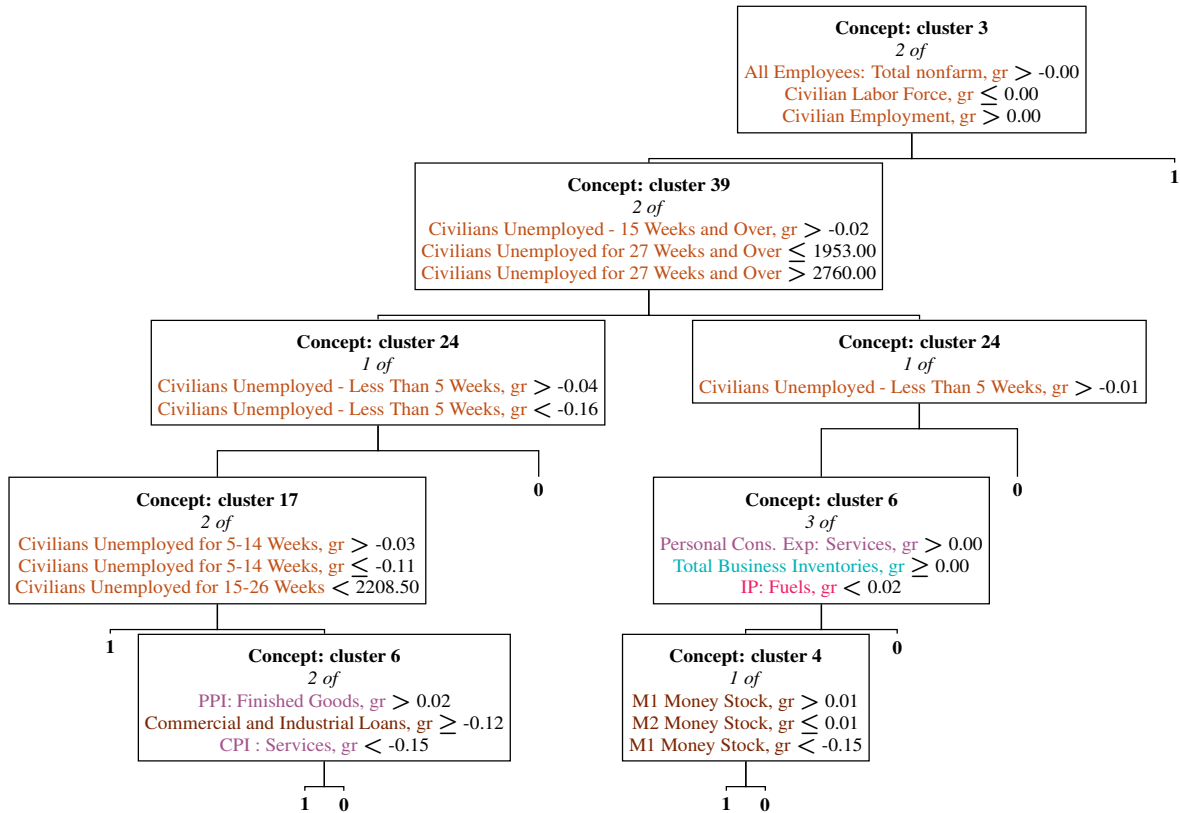
\begin{figure*}  
\centering

\resizebox*{16cm}{12cm}{
\subfigure[\textbf{Concept Tree-Expert. Variables from the same expert-defined concepts are displayed with the same color. Each node use variables from one single expert-based concept}]{ 
\input{tree_concept_expert.tex}
\label{fig:concept_expert}
}}

\resizebox*{16cm}{12cm}{
\subfigure[\textbf{Concept Tree-Correlated. Each node use variables from one single concept based on variable correlations. Variables from the same expert-defined concepts are displayed with the same color. }]{ 
\input{tree_concept_correlated.tex}
\label{fig:concept_correlated}
}}

\caption{\textbf{Structure of trained $3-of-3$ Decision Trees}}
\label{fig:trees}
\end{figure*}

\end{document}

%% file: illustration_tree_concept.tex
\tiny
\begin{tikzpicture}
\tikzset{every tree node/.style={align=center,anchor=north}}
\tikzset{level distance=60pt,sibling distance=5pt}
\tikzset{edge from parent/.style=
{draw,
edge from parent path={(\tikzparentnode.south)
-- +(0,-8pt)
-| (\tikzchildnode)}}}

\Tree[.\node[draw]{\\\textbf{Concept: Labor market}\\\textit{2 of }\\\textcolor{Bittersweet}{Civilians Unemployed - 15 Weeks and Over, gr} $\bm{>}$ -0.02\\\textcolor{Bittersweet}{Civilians Unemployed - Less Than 5 Weeks, gr} $\bm{\geq}$ -0.04\\\textcolor{Bittersweet}{Civilians Unemployed for 5-14 Weeks, gr} $\bm{>}$ 0.06};
	[.\node[draw]{\\\textbf{Concept: Output and income}\\\textit{2 of }\\\textcolor{WildStrawberry}{IP: Final Products and Nonindustrial Supplies, gr} $\bm{>}$ -0.00\\\textcolor{WildStrawberry}{IP: Nondurable Consumer Goods, gr} $\bm{\leq}$ -0.02\\\textcolor{WildStrawberry}{Real Personal Income, gr} $\bm{>}$ 0.00};
		\textbf{0}
		\textbf{1}
	]
	[.\node[draw]{\\\textbf{Concept: Consumption; orders; and inventories}\\\textit{2 of }\\\textcolor{BlueGreen}{Real Manu. and Trade Industries Sales, gr} $\bm{>}$ -0.02\\\textcolor{BlueGreen}{Unfilled Orders for Durable Goods, gr} $\bm{\geq}$ -0.01\\\textcolor{BlueGreen}{Total Business: Inventories to Sales Ratio, gr} $\bm{<}$ -0.01};
		\textbf{0}
		\textbf{1}
	]
]

\end{tikzpicture}


%% file: tree_trepan.tex

\tiny
\begin{tikzpicture}
\tikzset{every tree node/.style={align=center,anchor=north}}
\tikzset{level distance=50pt,sibling distance=5pt}
\tikzset{edge from parent/.style=
{draw,
edge from parent path={(\tikzparentnode.south)
-- +(0,-8pt)
-| (\tikzchildnode)}}}
\Tree[.\node[draw]{\\\textit{2 of }\\\textcolor{Bittersweet}{Civilians Unemployed - 15 Weeks and Over, gr} $\bm{>}$ -0.02\\\textcolor{Bittersweet}{Civilians Unemployed - Less Than 5 Weeks, gr} $\bm{\geq}$ -0.04\\\textcolor{Bittersweet}{Civilians Unemployed for 5-14 Weeks, gr} $\bm{>}$ 0.06};
	[.\node[draw]{\\\textit{2 of }\\\textcolor{Bittersweet}{Civilian Employment, gr} $\bm{>}$ -0.00\\\textcolor{Bittersweet}{Civilian Labor Force, gr} $\bm{\leq}$ 0.00\\\textcolor{BlueGreen}{Retail and Food Services Sales, gr} $\bm{>}$ 0.02};
		[.\node[draw]{\\\textit{3 of }\\\textcolor{WildStrawberry}{IP: Final Products and Nonindustrial Supplies, gr} $\bm{>}$ -0.00\\\textcolor{BurntOrange}{5-Year Treasury Rate, gr} $\bm{\geq}$ -0.04\\\textcolor{BurntOrange}{Japan / U.S. Foreign Exchange Rate, gr} $\bm{<}$ 0.00};
			[.\node[draw]{\\\textit{1 of }\\\textcolor{WildStrawberry}{Real personal income ex transfer receipts, gr} $\bm{>}$ 0.01\\\textcolor{black}{VXO} $\bm{\leq}$ 10.80\\\textcolor{Bittersweet}{Ratio of Help Wanted/No. Unemployed} $\bm{>}$ 1.08};
				\textbf{1}
				\textbf{0}
			]
			\textbf{1}
		]
		\textbf{1}
	]
	[.\node[draw]{\\\textit{3 of }\\\textcolor{Bittersweet}{All Employees: Total nonfarm, gr} $\bm{>}$ -0.00\\\textcolor{Bittersweet}{Civilians Unemployed for 15-26 Weeks} $\bm{\geq}$ 239.00\\\textcolor{Bittersweet}{Avg Hourly Earnings : Manufacturing, gr} $\bm{>}$ -0.00};
		[.\node[draw]{\\\textit{3 of }\\\textcolor{Bittersweet}{Civilian Labor Force, gr} $\bm{>}$ 0.00\\\textcolor{Bittersweet}{Civilian Employment, gr} $\bm{\leq}$ 0.00\\\textcolor{Brown}{MZM Money Stock, gr} $\bm{>}$ 0.00};
			\textbf{1}
			[.\node[draw]{\\\textit{3 of }\\\textcolor{BurntOrange}{6-Month Treasury Bill:, gr} $\bm{>}$ -0.01\\\textcolor{BlueGreen}{Housing Starts: Total New Privately Owned} $\bm{\leq}$ 2033.00\\\textcolor{BurntOrange}{Switzerland / U.S. Foreign Exchange Rate, gr} $\bm{>}$ -0.01};
				\textbf{0}
				[.\node[draw]{\\\textit{3 of }\\\textcolor{BurntOrange}{3-Month Treasury Bill:, gr} $\bm{>}$ -0.10\\\textcolor{WildStrawberry}{IP: Durable Materials, gr} $\bm{\leq}$ 0.02\\\textcolor{Brown}{Real Estate Loans at All Commercial Banks, gr} $\bm{<}$ 0.02};
					[.\node[draw]{\\\textit{1 of }\\\textcolor{WildStrawberry}{IP: Consumer Goods, gr} $\bm{>}$ 0.00\\\textcolor{WildStrawberry}{IP: Final Products and Nonindustrial Supplies, gr} $\bm{\leq}$ -0.00\\\textcolor{WildStrawberry}{IP: Fuels, gr} $\bm{>}$ 0.01};
						\textbf{0}
						\textbf{1}
					]
					\textbf{0}
				]
			]
		]
		\textbf{1}
	]
]
\end{tikzpicture}

%% file: tree_concept_expert.tex
\tiny
\begin{tikzpicture}
\tikzset{every tree node/.style={align=center,anchor=north}}
\tikzset{level distance=50pt,sibling distance=5pt}
\tikzset{edge from parent/.style=
{draw,
edge from parent path={(\tikzparentnode.south)
-- +(0,-8pt)
-| (\tikzchildnode)}}}
\Tree[.\node[draw]{\\\textbf{Concept: Labor market}\\\textit{2 of }\\\textcolor{Bittersweet}{Civilians Unemployed - 15 Weeks and Over, gr} $\bm{>}$ -0.02\\\textcolor{Bittersweet}{Civilians Unemployed - Less Than 5 Weeks, gr} $\bm{\geq}$ -0.04\\\textcolor{Bittersweet}{Civilians Unemployed for 5-14 Weeks, gr} $\bm{>}$ 0.06};
	[.\node[draw]{\\\textbf{Concept: Output and income}\\\textit{2 of }\\\textcolor{WildStrawberry}{IP: Final Products and Nonindustrial Supplies, gr} $\bm{>}$ -0.00\\\textcolor{WildStrawberry}{IP: Nondurable Consumer Goods, gr} $\bm{\leq}$ -0.02\\\textcolor{WildStrawberry}{Real Personal Income, gr} $\bm{>}$ 0.00};
		[.\node[draw]{\\\textbf{Concept: Interest and exchange rates}\\\textit{3 of }\\\textcolor{BurntOrange}{Japan / U.S. Foreign Exchange Rate, gr} $\bm{>}$ 0.00\\\textcolor{BurntOrange}{Moody’s Baa Corporate Bond Minus FEDFUNDS} $\bm{\leq}$ 4.49\\\textcolor{BurntOrange}{Japan / U.S. Foreign Exchange Rate, gr} $\bm{<}$ 0.03};
			\textbf{1}
			[.\node[draw]{\\\textbf{Concept: Stock market}\\\textit{3 of }\\\textcolor{CarnationPink}{SandP’s Composite Common Stock: Dividend Yield, gr} $\bm{>}$ -0.04\\\textcolor{CarnationPink}{VXO} $\bm{\geq}$ 10.77\\\textcolor{CarnationPink}{SandP’s Common Stock Price Index: Industrials, gr} $\bm{>}$ -0.04};
				[.\node[draw]{\\\textbf{Concept: Prices}\\\textit{2 of }\\\textcolor{DarkOrchid}{CPI : All Items, gr} $\bm{>}$ 0.00\\\textcolor{DarkOrchid}{Personal Cons. Exp: Nondurable goods, gr} $\bm{\leq}$ 0.01\\\textcolor{DarkOrchid}{PPI: Crude Materials, gr} $\bm{<}$ -0.03};
					\textbf{1}
					[.\node[draw]{\\\textbf{Concept: Money and credit}\\\textit{3 of }\\\textcolor{Brown}{Reserves Of Depository Institutions, gr} $\bm{>}$ -0.04\\\textcolor{Brown}{Total Reserves of Depository Institutions, gr} $\bm{\leq}$ 0.06\\\textcolor{Brown}{M1 Money Stock, gr} $\bm{>}$ -0.10};
						\textbf{0}
						\textbf{1}
					]
				]
				\textbf{1}
			]
		]
		\textbf{1}
	]
	[.\node[draw]{\\\textbf{Concept: Consumption; orders; and inventories}\\\textit{2 of }\\\textcolor{BlueGreen}{Real Manu. and Trade Industries Sales, gr} $\bm{>}$ -0.02\\\textcolor{BlueGreen}{Unfilled Orders for Durable Goods, gr} $\bm{\geq}$ -0.01\\\textcolor{BlueGreen}{Total Business: Inventories to Sales Ratio, gr} $\bm{<}$ -0.01};
		[.\node[draw]{\\\textbf{Concept: Prices}\\\textit{1 of }\\\textcolor{DarkOrchid}{Personal Cons. Exp: Services, gr} $\bm{>}$ 0.00\\\textcolor{DarkOrchid}{CPI : All items less shelter, gr} $\bm{\geq}$ 0.00\\\textcolor{DarkOrchid}{Personal Cons. Exp: Services, gr} $\bm{<}$ 0.00};
			[.\node[draw]{\\\textbf{Concept: Interest and exchange rates}\\\textit{2 of }\\\textcolor{BurntOrange}{10-Year Treasury Rate, gr} $\bm{>}$ 0.00\\\textcolor{BurntOrange}{Switzerland / U.S. Foreign Exchange Rate, gr} $\bm{\geq}$ 0.01\\\textcolor{BurntOrange}{Moody’s Seasoned Aaa Corporate Bond Yield, gr} $\bm{<}$ 0.04};
				\textbf{0}
				[.\node[draw]{\\\textbf{Concept: Output and income}\\\textit{3 of }\\\textcolor{WildStrawberry}{IP: Manufacturing (SIC), gr} $\bm{>}$ 0.01\\\textcolor{WildStrawberry}{IP: Durable Consumer Goods, gr} $\bm{\leq}$ 0.03\\\textcolor{WildStrawberry}{Real Personal Income, gr} $\bm{>}$ -0.00};
					\textbf{0}
					\textbf{1}
				]
			]
			\textbf{0}
		]
		\textbf{1}
	]
]
\end{tikzpicture}


%% file: tree_concept_correlated.tex

\tiny
\begin{tikzpicture}
\tikzset{every tree node/.style={align=center,anchor=north}}
\tikzset{level distance=50pt,sibling distance=5pt}
\tikzset{edge from parent/.style=
{draw,
edge from parent path={(\tikzparentnode.south)
-- +(0,-8pt)
-| (\tikzchildnode)}}}
\Tree[.\node[draw]{\\\textbf{Concept: cluster 3}\\\textit{2 of }\\\textcolor{Bittersweet}{All Employees: Total nonfarm, gr} $\bm{>}$ -0.00\\\textcolor{Bittersweet}{Civilian Labor Force, gr} $\bm{\leq}$ 0.00\\\textcolor{Bittersweet}{Civilian Employment, gr} $\bm{>}$ 0.00};
	[.\node[draw]{\\\textbf{Concept: cluster 39}\\\textit{2 of }\\\textcolor{Bittersweet}{Civilians Unemployed - 15 Weeks and Over, gr} $\bm{>}$ -0.02\\\textcolor{Bittersweet}{Civilians Unemployed for 27 Weeks and Over} $\bm{\leq}$ 1953.00\\\textcolor{Bittersweet}{Civilians Unemployed for 27 Weeks and Over} $\bm{>}$ 2760.00};
		[.\node[draw]{\\\textbf{Concept: cluster 24}\\\textit{1 of }\\\textcolor{Bittersweet}{Civilians Unemployed - Less Than 5 Weeks, gr} $\bm{>}$ -0.04\\\textcolor{Bittersweet}{Civilians Unemployed - Less Than 5 Weeks, gr} $\bm{<}$ -0.16};
			[.\node[draw]{\\\textbf{Concept: cluster 17}\\\textit{2 of }\\\textcolor{Bittersweet}{Civilians Unemployed for 5-14 Weeks, gr} $\bm{>}$ -0.03\\\textcolor{Bittersweet}{Civilians Unemployed for 5-14 Weeks, gr} $\bm{\leq}$ -0.11\\\textcolor{Bittersweet}{Civilians Unemployed for 15-26 Weeks} $\bm{<}$ 2208.50};
				\textbf{1}
				[.\node[draw]{\\\textbf{Concept: cluster 6}\\\textit{2 of }\\\textcolor{DarkOrchid}{PPI: Finished Goods, gr} $\bm{>}$ 0.02\\\textcolor{Brown}{Commercial and Industrial Loans, gr} $\bm{\geq}$ -0.12\\\textcolor{DarkOrchid}{CPI : Services, gr} $\bm{<}$ -0.15};
					\textbf{1}
					\textbf{0}
				]
			]
			\textbf{0}
		]
		[.\node[draw]{\\\textbf{Concept: cluster 24}\\\textit{1 of }\\\textcolor{Bittersweet}{Civilians Unemployed - Less Than 5 Weeks, gr} $\bm{>}$ -0.01};
			[.\node[draw]{\\\textbf{Concept: cluster 6}\\\textit{3 of }\\\textcolor{DarkOrchid}{Personal Cons. Exp: Services, gr} $\bm{>}$ 0.00\\\textcolor{BlueGreen}{Total Business Inventories, gr} $\bm{\geq}$ 0.00\\\textcolor{WildStrawberry}{IP: Fuels, gr} $\bm{<}$ 0.02};
				[.\node[draw]{\\\textbf{Concept: cluster 4}\\\textit{1 of }\\\textcolor{Brown}{M1 Money Stock, gr} $\bm{>}$ 0.01\\\textcolor{Brown}{M2 Money Stock, gr} $\bm{\leq}$ 0.01\\\textcolor{Brown}{M1 Money Stock, gr} $\bm{<}$ -0.15};
					\textbf{1}
					\textbf{0}
				]
				\textbf{0}
			]
			\textbf{0}
		]
	]
	\textbf{1}
]
\end{tikzpicture}